# Exploring LLM Autoscoring Reliability in Large-Scale Writing Assessments Using Generalizability Theory

Dan Song, University of Iowa

Won-Chan Lee, University of Iowa

Hong Jiao, University of Maryland

## Abstract

This study investigates the estimation of reliability for large language models (LLMs) in scoring writing tasks from the AP Chinese Language and Culture Exam. Using generalizability theory, the research evaluates and compares score consistency between human and AI raters across two types of AP Chinese free-response writing tasks: story narration and email response. These essays were independently scored by two trained human raters and seven AI raters. Each essay received four scores: one holistic score and three analytic scores corresponding to the domains of task completion, delivery, and language use. Results indicate that although human raters produced more reliable scores overall, LLMs demonstrated reasonable consistency under certain conditions, particularly for story narration tasks. Composite scoring that incorporates both human and AI raters improved reliability, which supports that hybrid scoring models may offer benefits for large-scale writing assessments.

**Exploring LLM Autoscoring Reliability in Large-Scale Writing Assessments Using**

**Generalizability Theory**

The integration of large language models (LLMs) into automated essay scoring (AES) represents a significant shift in how essay scoring is approached. While traditional AES systems have long depended on manually engineered features and statistical models (Attali & Burstein, 2006; Dikli, 2006), LLMs offer the potential to assess student writing with greater flexibility and contextual sensitivity by drawing on deep learning architectures trained on diverse textual corpora (Ifenthaler, 2022; Ouyang et al., 2022). However, despite their promising capabilities, recent studies indicate that LLMs have not yet consistently matched the scoring reliability of established AES tools or trained human raters, especially in high-stakes language assessment contexts (Mizumoto & Eguchi, 2023; Xiao et al., 2025; Yancey et al., 2023). These concerns highlight the need for rigorous evaluation of LLM-based scoring systems, particularly with respect to their reliability and alignment with human scoring standards. This study addresses these challenges by applying generalizability theory to systematically examine the consistency of LLM-generated scores on standardized writing tasks in the AP Chinese Language and Culture Exam (AP Chinese Exam).

## Literature Review

This section reviews the literature on AES and the application of LLMs to AES. It also provides brief overviews of generalizability theory and the AP Chinese Language and Culture Exam, followed by the research questions addressed in this study.

### Automated Essay Scoring

The field of AES has evolved significantly since its inception in the 1960s, leveraging advancements in natural language processing (NLP) and machine learning to evaluate student writing (Chen & He, 2013). Initially, essay scoring relied entirely on human raters, which made the grading process both time-consuming and often inconsistent across raters. In large-scale standardized tests such as the Graduate Record Examination (GRE), essays were traditionally rated by two human evaluators to mitigate individual rating errors. However, with the development of AES systems, automated tools have been increasingly integrated into writing assessments, often serving as a secondary rater to improve efficiency, consistency, and reliability (Dikli, 2006).

AES systems aim to replicate human scores by utilizing a combination of statistical modeling, machine learning, and NLP techniques. These models are typically trained on extensive datasets of human-scored essays, allowing them to learn patterns that align with human scoring standards (Ifenthaler, 2022). The first and one of the best AES models, Project Essay Grade (PEG), introduced by Page (1966, 1968), used simple regression models to estimate scores based on text features such as essay length and word frequency. Later, more models emerged, including Pearson's Intelligent Essay Assessor (Landauer et al., 2003), Vantage Learning's IntelliMetric (Elliott, 2003), and Educational Testing Service's e-Rater (Attali & Burstein, 2006; Burstein, 2003). These AES systems incorporated different techniques to assess key writing characteristics such as grammar, coherence, and organization.

Common AES models typically operate through a two-step process. In the first step, NLP techniques extract linguistic features (such as lexical, syntactic, semantic, rhetorical features) relevant to writing quality. In the second step, statistical or machine learning algorithms predict numeric scores based on patterns observed from the extracted features (McCarthy et al., 2022; Strobl et al., 2019). While early AES tools relied heavily on handcrafted feature engineering, newer systems increasingly incorporate deep learning methods to enhance scoring accuracy.

Despite these advancements, automated scoring still faces limitations, particularly in high-stakes assessments such as second-language (L2) writing evaluations. Human scoring remains the gold standard due to its understanding of complex linguistic features, despite challenges such as high costs, rater fatigue, and subjectivity (Zhao et al., 2023). The emergence of LLMs has introduced new possibilities for improving AES by offering more sophisticated text analysis capabilities.

**LLMs for Automated Essay Scoring**

Since the release of ChatGPT 3.5 in November 2022, LLMs have gained attention as tools for automated essay evaluation. Unlike traditional AES systems, which rely on extracted linguistic features, LLMs leverage deep learning architectures to process language more holistically, drawing on extensive pre-training over large text corpora. These models can generate human-like responses, follow complex instructions, and adapt to diverse writing prompts, which positions them as potential alternatives to conventional AES approaches (Ouyang et al., 2022).

Despite these advantages, LLMs have not consistently outperformed traditional AES models or human raters. For instance, Mizumoto and Eguchi (2023) examined ChatGPT's ability to score essays using the text-davinci-003 model and incorporated the IELTS scoring rubrics into their prompts. The rubrics provided a detailed evaluation of (1) task response, (2) coherence and cohesion, (3) lexical resource, and (4) grammatical range and accuracy, using a 10-score scale ranging from 0 to 9. While AES using GPT can achieve a certain level of accuracy, it still falls short of full agreement with human raters and should therefore be used alongside human evaluation. Similarly, Yancey et al. (2023) evaluated GPT-4's few-shot capabilities in predicting Common European Framework of Reference for Languages levels for short essays written by second-language learners. Their findings showed that the model's Quadratic Weighted Kappa (QWK) scores were lower than those of traditional machine learning models such as XGBoost and fell short of human raters' performance. Further, Xiao et al. (2025) investigated GPT-4's zero-shot and few-shot performance on the Automated Student Assessment Prize dataset, revealing the model's struggle to achieve high QWK scores, raising concerns about its consistency in automated scoring tasks.

To address these limitations, researchers have explored structured prompting frameworks to refine LLM-based scoring (Han et al., 2023; Stahl et al., 2024). However, these approaches have yet to consistently surpass existing AES models. A key challenge lies in prompt engineering—how instructions and contextual cues are framed for the model. While LLMs exhibit strong reasoning abilities, assigning scores based on established scoring rubrics remains challenging.

Meanwhile, while prompt engineering remains a central challenge in refining LLM-based scoring, another approach via fine-tuning LLMs on specific scoring tasks has also shown potential to improve accuracy. However, it requires considerable computational resources and large annotated datasets, which makes it impractical for many educational settings. The variability in essay prompts, grading rubrics, and assessment expectations across institutions further complicates large-scale implementation. Instead of fine-tuning for every context, researchers are exploring AI architectures that mimic human cognitive processes, particularly in reasoning and planning (Fountas et al., 2024). These models aim to enhance the interpretative depth and consistency of LLM-based scoring, which allows for more reliable evaluations without extensive retraining.

Given the costs and time constraints associated with fine-tuning, research on zero-shot and few-shot learning continues to attract attention. While LLMs offer scalable and flexible AES solutions, concerns persist regarding their reliability and alignment with human raters. Reliability and validity remain fundamental to the quality of any assessment (Hyland, 2003).

Reliability refers to the consistency of test takers' scores across different occasions, tasks, or raters (Bachman, 1990). Scoring reliability is particularly critical in writing assessment due to the complexity of evaluating writing performance and the potential for rater bias. Score inconsistencies arise from variability in rater judgments, rubric types (holistic vs. analytic), and student factors such as language proficiency and conceptual knowledge (Baker, 2010). Additionally, rater subjectivity arising from differences in rating behaviors, decision-making processes, and scoring tendencies remains a significant source of measurement error. Ensuring fairness in high-stakes assessments requires both inter-rater reliability (agreement among different raters) and intra-rater reliability (consistency within a single rater over time). While rater training can help mitigate inconsistencies, it is often limited by time and cost constraints (Weigle, 2002).

Recent research has explored the role of AI in essay evaluation, which highlights both its strengths and limitations. While human raters excel in providing comprehensive feedback, AI-driven tools such as ChatGPT show promise in automated scoring for educational assessment (Latif & Zhai, 2024; Steiss et al., 2024). Additionally, AI models fine-tuned for specific domains have demonstrated high accuracy in assessing educational data, which suggests their potential for targeted writing assessments (Latif & Zhai, 2024). However, ensuring their consistency remains an ongoing challenge, which underscores the importance for continued in-depth research in better understanding LLM-based scoring and variances in scores assigned by LLM raters.

**Generalizability Theory**

Generalizability (G) theory provides a comprehensive framework for assessing the reliability of a measurement system by systematically analyzing multiple sources of measurement error. Unlike classical test theory, which assumes a single undifferentiated error term, G theory decomposes measurement error into distinct variance components such as raters, test items, or testing occasions using ANOVA-based techniques. This approach allows researchers to quantify the relative contributions of different error sources, offering deeper insights into measurement precision and reliability.

Additionally, G theory is particularly valuable for evaluating the dependability of measurement procedures and optimizing assessment designs (Brennan, 2001; Cronbach et al., 1972). Through a G study, researchers can estimate how much variability is introduced by different facets of the measurement process. Based on these findings, Decision (D) studies can then be conducted to determine the optimal measurement conditions that minimize error and enhance score reliability. The ability to distinguish and model multiple sources of error makes G theory a powerful tool for improving the interpretability of test scores across diverse assessment contexts.

Several studies have applied G theory to AES in educational contexts. Wilson et al. (2019) and Chen, Hebert, and Wilson (2022) used G theory to examine the reliability of scores from PEG in elementary writing tasks. Their findings showed that AES could achieve reliability levels comparable to human raters for non-struggling writers, but struggling writers required more tasks or ratings to reach acceptable reliability. Moreover, their studies identified task and genre as significant sources of variance and highlighted the benefits of combining human and AES ratings to enhance both reliability and efficiency. Similarly, Wilson et al. (2018)

demonstrated that AES could support low- and high-stakes decisions in Response to Intervention (RTI) frameworks, provided that sufficient tasks and genres were administered.

Building on these findings, the present study applies G theory to examine the reliability of scores from human and LLM-based raters in a high-stakes (i., e., AP Chinese Language Culture Exam), second-language exam. By modeling multiple sources of error variance, including rater type (human vs. AI), task type (story narration vs. email response), and scoring domain (task completion, delivery, language use), this study extends prior work on traditional AES to a novel setting involving LLMs and high-stakes language assessment.

**AP Chinese Language Culture Exam**

The AP Chinese Exam is a standardized assessment administered by the College Board as part of the Advanced Placement (AP) program in the United States. The exam evaluates skills in interpersonal, interpretive, and presentational communication, while incorporating cultural knowledge. The test consists of multiple-choice questions assessing reading and listening comprehension, as well as free-response tasks that require spoken and written responses. It is unique among AP language exams for its computer-based format, which includes a built-in input system for typing in simplified or traditional Chinese characters and recording spoken responses. The exam aligns with college-level Chinese language courses, allowing students to earn college credit or advanced placement depending on their scores. By assessing both linguistic and cultural competencies, the AP Chinese Exam encourages students to develop practical communication skills while deepening their understanding of Chinese traditions, contemporary society, and global connections.

**Purpose of the Study**

The purpose of the study is to use G theory to systematically assess how different essay writing facets and their interactions affect reliability, and to model how reliability might improve under alternative scoring designs. This research utilizes writing samples collected from an experimental AP Chinese Exam, with each sample evaluated by 2 human raters and 7 AI raters. In this study, G theory was chosen for its flexibility and depth needed to evaluate the complex sources of measurement error that arise when comparing human and LLM-based raters across multiple tasks and scoring domains. Unlike classical reliability indices such as Cronbach's alpha or inter-rater correlations, which assume a single undifferentiated error term, G theory can separate the contributions of various facets, such as raters, tasks, and their interactions, to overall score variability. This is particularly important in this research because the study design involves not only different types of raters (human and AI) but also distinct task types (story narration and email response) and multiple scoring dimensions (task completion, delivery, and language use). Moreover, recent studies (Wilson, Chen, Sandbank, & Hebert, 2019; Chen, Hebert, & Wilson, 2022) have demonstrated the utility of G theory in AES contexts, providing a strong conceptual and methodological precedent for applying it to LLM-based scoring evaluation. This makes G theory especially well-suited to meet the objectives of this study.

This study is guided by the following research questions:

1. How reliable are scores produced by LLMs in scoring large-scale writing assessments compared to human raters?

2. How do LLMs and human raters differ in consistency across task types: email responses and story narration?

3. How do LLMs and human raters differ in evaluating the domains of task completion, delivery, and language use in AP Chinese exam writing tasks?

4. Can composite scoring (human + AI raters) improve the overall reliability of writing assessments?

## Methods

To address the research questions, multivariate generalizability theory (MGT) was employed, with each question involving a distinct design structure. The MGT approach offers a flexible framework for quantifying measurement error and reliability, particularly when the scores of interests are composites across levels of a fixed facet. For each research question, a corresponding G study was conducted, followed by a series of D studies to explore various sample size conditions for the relevant facets.

### Participants

The study consists of 30 students enrolled in third-year or higher Mandarin courses at U.S. colleges, all of whom are non-native learners of Mandarin. Their ages range from 20 to 25 years. Sixteen students are in their third year, and 14 are in their fourth year. The participants come from diverse academic backgrounds, with majors in fields such as business, the humanities, STEM, and international relations. According to the College Board (2024), the AP Chinese Language and Culture course corresponds to the proficiency level of a fourth-semester university Mandarin course. Based on this equivalency, we consider that students who have completed at least two years of college-level Mandarin possess the necessary language skills to complete the AP Chinese exam's free-response writing tasks.

### Data Collection

This study analyzes the writing section of the free-response portion from the 2021 and 2022 assessments, which consists of four tasks: two story narration (SN) tasks and two email response (ER) tasks. In the SN tasks, students are asked to create a narrative based on four sequential images within a 15-minute time limit. In the ER tasks, students respond to an email message, also within a 15-minute timeframe. A total of 30 student writing samples were included in the study.

A total of 120 essays were independently scored by two human AP Chinese raters and seven AI raters. The human raters, both are AP Chinese and Culture Exam raters, assessed the writing samples using the official AP Chinese Language and Culture Scoring Guidelines established by the College Board. The scoring framework focused on three key dimensions: (a) task completion, (b) delivery, and (c) language use. Raters began with a holistic assessment of each essay, then assigned analytic scores for each domain. As a result, each participant received four scores per essay: one overall holistic score and three domain-specific analytic scores, each ranging from 0 to 6 (Song & Tang, 2025).

Beginning in October 2023, ChatGPT-3.5 was used to score the same 120 essays. Later, we intentionally selected seven AI engines to represent a diverse range of architectures and capabilities of the latest development in LLMs. The AI raters included ChatGPT-3.5, ChatGPT-4.0, ChatGPT-4o, ChatGPTo1, Gemini 1.5, Gemini 2.0, and Claude 3.5 Sonnet. Data collection and AI training extended over approximately 18 months.

Following the recommendations of Yancey et al. (2023), the training emphasized the importance of clear prompts, well-defined criteria, and appropriate calibration based on the complexity of student texts. In line with Lee et al. (2024), their analysis comparing zero-shot and few-shot learning prompts demonstrated a notable improvement in scoring accuracy when few-shot learning was applied. In this study, before AI scoring, the AI models were trained using task-specific protocols designed for each of the four writing tasks: SN1, SN2, ER1, and ER2. These training documents included detailed writing prompts, scoring rubrics, and sample responses benchmarked across multiple score levels. To ensure the validity of AI scoring, representative examples were selected from the College Board website and previously human scored responses, as part of the AI training protocol for grading student samples.

AI raters were trained to follow the same scoring framework as human raters, prioritizing task completion first, followed by delivery and language use, in alignment with official AP rubrics. For SN tasks, addressing all four pictures was necessary to earn a top score; for ER tasks, students had to respond to all questions in the email to fulfill the task. A full training document on SN2: AI Training Protocol for Grading Student Samples is provided in Appendix A as a representative example.

## Results

In this paper, the results are presented alongside the G theory analyses to highlight the close connection between the methodological design and the findings. We organized the results section in this way to provide readers with an integrated view of how the G theory framework informs and contextualizes the reliability estimates, variance components, and comparisons of human and AI raters. This structure allows for a more coherent discussion of both the analytical approach and its outcomes, which is particularly helpful given the multifaceted nature of the study.

Before presenting the G theory analyses, Appendix B summarizes the descriptive statistics (means and standard deviations) of scores assigned by human and AI raters across the four writing tasks and scoring domains. These results provide an initial view of scoring patterns, showing that human raters tended to assign higher and more consistent scores compared to AI raters, while AI raters exhibited greater variability across tasks and domains. This descriptive overview sets the stage for the subsequent G theory analyses, which quantify and separate the sources of variability in these scores.

**Generalizability Theory Analysis**

To address the first three research questions, MGT analyses were conducted. Let $p$ represent the object of measurement (i.e., persons), while $t$ and $r$ denote the measurement facets—tasks and raters, respectively. Here, both $t$ and $r$ are treated as *random* facets because the specific tasks and raters included in the study are considered representative samples drawn from larger universes of possible tasks and raters, and the goal is to generalize findings beyond the particular tasks and raters used in this study. The two task types, SN and ER, serve as a fixed multivariate facet, denoted as $v$. The multivariate G study design is represented as $p^{\bullet} \times t^{\circ} \times r^{\bullet}$, where both $p$ and $r$ are crossed with the two levels of the multivariate facet (task types), indicated by a bullet, while $t$ is nested within task types, indicated by an open circle. This design implies that in repeated measurement procedures, each form will always include the same two task types (SN and ER) with different sets of prompts, while raters are assumed to vary across replications.

The left panel of Figure 1 presents a Venn diagram representation of the G study design. The dashed circle denotes the multivariate facet ($v$), while the three solid circles represent the main effects ($p$, $t$, and $r$), with their intersections indicating interaction effects. The seven distinct regions formed by these solid circles correspond to the following score effects: the main effects ($p$, $t$, and $r$); their two-way interactions ($pt$, $pr$, and $tr$); and the three-way interaction ($ptr$), which includes residual variance. Variance components for these seven score effects are estimated separately for each level of $v$. Because $p$, $r$, and their interaction ($pr$) are crossed with $v$, these effects also include covariance terms.

The multivariate design analysis was conducted separately for human and AI raters and was repeated four times for the overall holistic score and three domain-specific scores. This paper presents a subset of the results.

To address the fourth research question, a different design structure was used, with rater type (human vs. AI) serving as the fixed multivariate facet ($v$), while raters ($r$) and tasks ($t$) were treated as random facets. The design is represented as $p^\bullet \times r^\circ \times t^\bullet$ (see the right Venn Diagram in Figure 1). Separate analyses were conducted for the SN and ER task types.

The G study data were collected with the following sample sizes: $n_p = 30$, $n_t = 2$ for each task type, $n_r = 2$ human raters, and $n_r = 7$ AI raters. These data were used to estimate G study variance and covariance components, which served as the basis for various D studies.

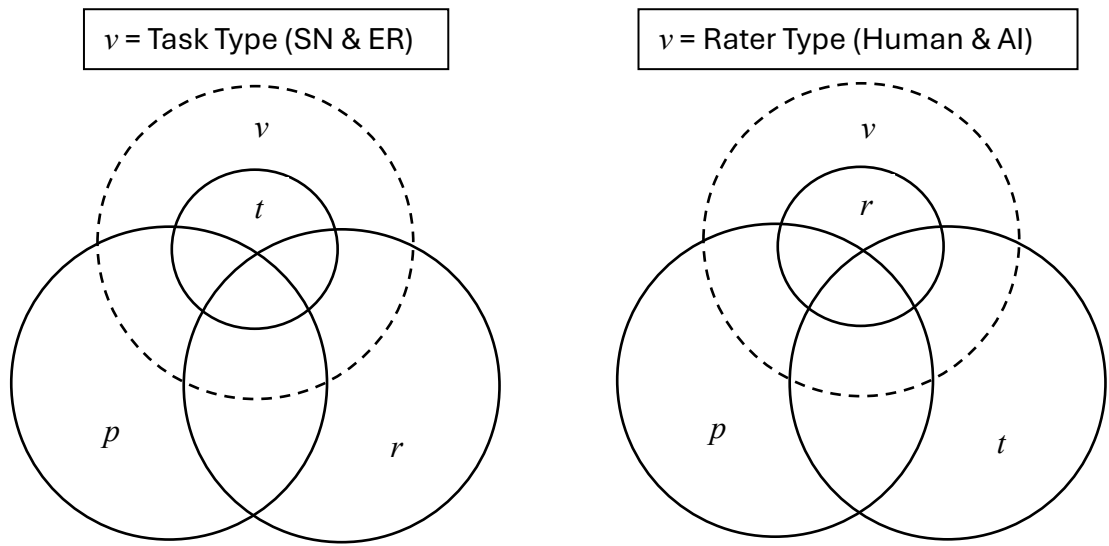

*Figure 1.* Venn diagram representations of $p^\bullet \times t^\circ \times r^\bullet$ and $p^\bullet \times r^\circ \times t^\bullet$ designs.

## Comparison of Human and AI Raters (Research Question 1)

To address this question, analyses focused on the overall holistic scores. G study variance and covariance components were estimated for the seven score effects associated with the $p^{\bullet} \times t^{\circ} \times r^{\bullet}$ design. Table 1 presents these estimates, providing a comparison between human and AI raters. Negative variance estimates arising from sampling error were replaced with zero.

Table 1

*G-Study Variance and Covariance Components: Overall Scores*

| | Human Rater | | | AI Rater | |
|---|---|---|---|---|---|
| | **SN** | **ER** | | **SN** | **ER** |
| $\Sigma_p =$ | $\begin{bmatrix} 0.286 & 0.302 \\ 0.302 & 0.504 \end{bmatrix}$ | | $\Sigma_p =$ | $\begin{bmatrix} 0.345 & 0.280 \\ 0.280 & 0.345 \end{bmatrix}$ | |
| $\Sigma_t =$ | $\begin{bmatrix} 0.090 & \\ & 0.103 \end{bmatrix}$ | | $\Sigma_t =$ | $\begin{bmatrix} 0.037 & \\ & 0.000 \end{bmatrix}$ | |
| $\Sigma_r =$ | $\begin{bmatrix} 0.091 & 0.097 \\ 0.097 & 0.096 \end{bmatrix}$ | | $\Sigma_r =$ | $\begin{bmatrix} 0.036 & 0.058 \\ 0.058 & 0.000 \end{bmatrix}$ | |
| $\Sigma_{pt} =$ | $\begin{bmatrix} 0.093 & \\ & 0.297 \end{bmatrix}$ | | $\Sigma_{pt} =$ | $\begin{bmatrix} 0.130 & \\ & 0.433 \end{bmatrix}$ | |
| $\Sigma_{pr} =$ | $\begin{bmatrix} 0.059 & 0.028 \\ 0.028 & 0.054 \end{bmatrix}$ | | $\Sigma_{pr} =$ | $\begin{bmatrix} 0.094 & 0.071 \\ 0.071 & 0.021 \end{bmatrix}$ | |
| $\Sigma_{tr} =$ | $\begin{bmatrix} 0.000 & \\ & 0.004 \end{bmatrix}$ | | $\Sigma_{tr} =$ | $\begin{bmatrix} 0.070 & \\ & 0.188 \end{bmatrix}$ | |
| $\Sigma_{ptr} =$ | $\begin{bmatrix} 0.068 & \\ & 0.088 \end{bmatrix}$ | | $\Sigma_{ptr} =$ | $\begin{bmatrix} 0.311 & \\ & 0.396 \end{bmatrix}$ | |

Each of the seven variance-covariance matrices contains two columns corresponding to the two task types. Several key observations emerge. First, task variability, represented by $\Sigma_t$ indicates that prompts differ more in average difficulty with human raters than with AI raters for both SN and ER task types (i.e., .090 > .013 and .103 > .000). Second, human raters exhibit greater variability in average stringency than AI raters for both SN and ER, as reflected in the diagonal elements of $\Sigma_r$. However, the lower variability among AI raters does not necessarily translate to higher reliability of AI scores. Reliability, in the traditional sense, such as coefficient alpha, concerns the consistency of rank ordering examinee performance over repeated measurements. The interaction terms involving $p$, $\Sigma_{pt}$, $\Sigma_{pr}$, and $\Sigma_{ptr}$ indicate that larger variances are associated with AI raters compared to human raters. This suggests that AI raters exhibit greater variation in the relative ordering of persons, which contributes to lower reliability of AI-rated scores, as discussed later in the D studies.

Various D studies were conducted to estimate the reliability of ratings using several sample sizes for tasks and raters. While the variance and covariance components from the G study shown in Table 1 are for *single* persons, raters, and tasks, the focus of a D study is on the variance and covariance of *mean* scores over the specified set of measurement conditions. The primary outcomes of a D study include the universe score variance (analogous to true score variance in classical test theory), reliability-like coefficients, and error variances, among others. In this analysis, we primarily focus on the generalizability coefficient, which is analogous to the classical reliability coefficient.

For the comparison between human and AI raters, D study results were obtained using the *mean* overall holistic scores across both SN and ER task types. The D study employed the same design as the G study, denoted as $p^\bullet \times T^\circ \times R^\bullet$, where the uppercase letters $T$ and $R$ emphasize that the D study focuses on mean scores. The following D study sample sizes were considered: $n_t' = 1, 2, 3, 4$ for each task type and $n_r' = 1, 2, 3, 4$. For example, a D study with $n_t' = 1$ and $n_r' = 2$ represents a measurement procedure where each individual responds to one SN task and one ER task, both rated by the same two raters. In total, 16 D studies were conducted.

Figure 2 summarizes the generalizability coefficients obtained from the 16 D studies. The X-axis represents the number of tasks ($n_t'$) ranging from 1 to 4, while the four lines correspond to the number of raters ($n_r'$) from 1 to 4. The solid line at 0.8 serves as a benchmark for assessing whether the reliability meets the investigator's desired threshold.

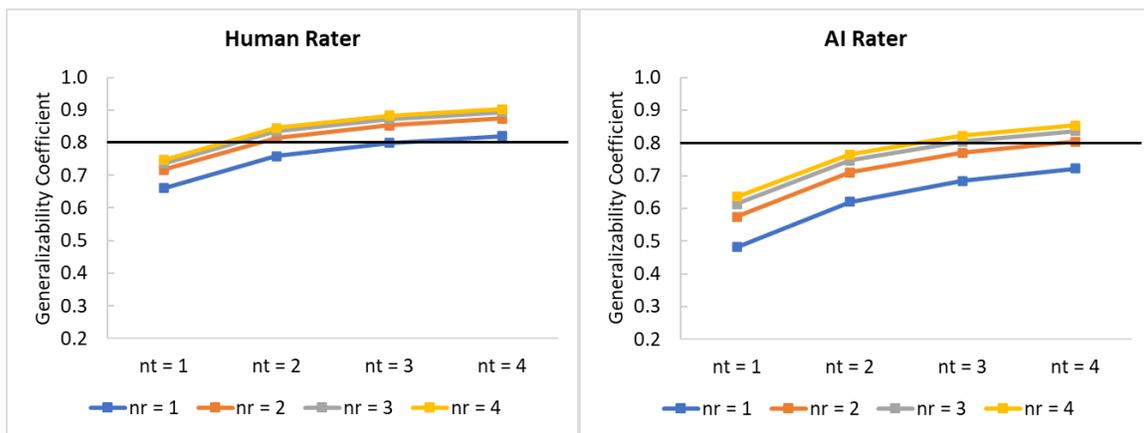

*Figure 2.* Generalizability coefficients: Comparison of human and AI raters.

The results in the left plot, which report findings for human raters, indicate that generalizability coefficients were consistently higher than those for AI raters, as shown in the right plot, under the same D study sample-size conditions. For example, with $n_t' = 2$ and $n_r' = 2$, the generalizability coefficient was 0.81 for human raters, compared to 0.71 for AI raters.

For both human and AI raters, the generalizability coefficient increased as the number of tasks increased, reinforcing that more tasks lead to more reliable scores. Similarly, increasing the number of raters improved reliability, with the largest gain observed when going from 1 to 2 raters. However, the improvement from 2 to 3 raters was minimal, suggesting that using more than two raters may not be necessary. Focusing on human raters, when only one human rater is used (represented by the blue line), at least four prompts per task type are required to achieve the target reliability. In contrast, with two raters, only two prompts per task type would suffice.

### *Comparison of Task Types (Research Question 2)*

The overall holistic scores are again used to compare the results for the two task types, SN and ER. The G study variance and covariance components for each of the SN and ER task types are reported in Table 1. The two columns of variance components for SN and ER can also be obtained through separate univariate analyses using a $p \times t \times r$ design. In general, for both human and AI raters, the variance components were similar between SN and ER, except for the $pt$ interaction, where ER exhibited substantially larger variance compared to SN. This difference contributes to the error variance and generalizability coefficients observed in the D studies.

D study results for generalizability coefficients are presented in the plots in Figure 3. The two plots in the first-row show results for SN for human and AI raters, while the second row displays ER results. In general, SN scores were more reliable than ER scores, with the differences being more pronounced with AI raters. For example, the difference in generalizability coefficients between SN and ER for human raters with $n_r' = 2$ and $n_t' = 2$ was $.755 - .719 = .036$; by contrast, for AI raters, it was $.646 - .514 = .132$.

The increase in reliability with the number of raters was smaller for ER than for SN, and this trend was consistent for both human and AI raters. In particular, using more than two human raters for the ER task type had minimal impact on reliability.

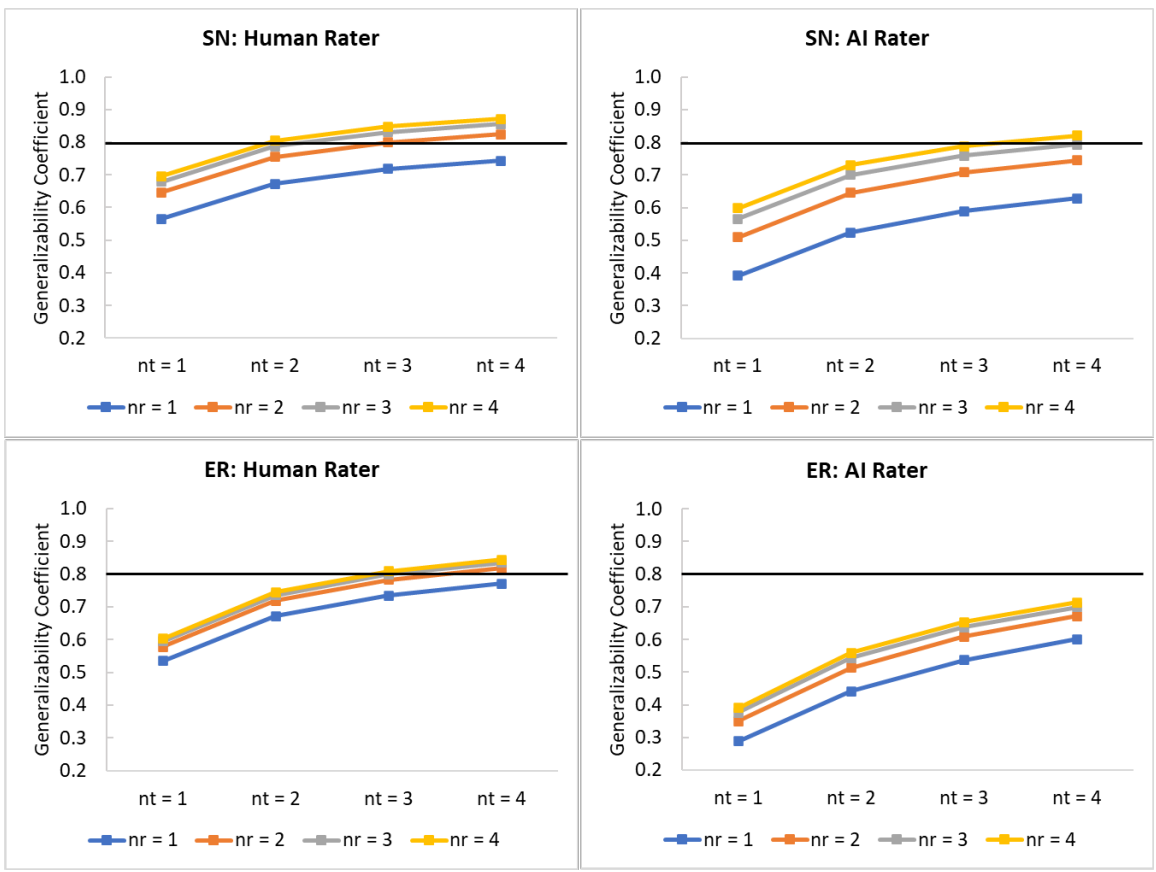

*Figure 3.* Generalizability coefficients: Comparison of SN and ER task types.

### Comparison Across Domain-Specific Scores (Research Question 3)

As previously mentioned, in addition to the overall holistic scores, three domain-specific scores were generated: task completion (TC), delivery (DL), and language use (LU). A MGT analysis using the $p^{\bullet} \times t^{\circ} \times r^{\bullet}$ design was conducted for each domain score, and the estimated G study variance and covariance components are presented in Tables 2 and 3 for human and AI raters, respectively.

In general, the relative magnitudes of the variance components were similar across the three domain scores for both human and AI raters, with the $pt$ interaction showing the largest variance. The variance associated with the object of measurement was highest for TC, which corresponds to the universe score variance in D studies. Consequently, TC is expected to exhibit relatively higher reliability compared to the other two domain scores.

Table 2

*G-Study Variance and Covariance Components: Human Rater*

| Task Completion | | | Delivery | | | Language Use | | |
|---|---|---|---|---|---|---|---|---|
| | **SN** | **ER** | | **SN** | **ER** | | **SN** | **ER** |
| $\Sigma_p =$ | $\begin{bmatrix} 0.328 & 0.458 \\ 0.458 & 0.794 \end{bmatrix}$ | | $\Sigma_p =$ | $\begin{bmatrix} 0.245 & 0.323 \\ 0.323 & 0.549 \end{bmatrix}$ | | $\Sigma_p =$ | $\begin{bmatrix} 0.241 & 0.263 \\ 0.263 & 0.526 \end{bmatrix}$ | |
| $\Sigma_t =$ | $\begin{bmatrix} 0.080 & \\ & 0.000 \end{bmatrix}$ | | $\Sigma_t =$ | $\begin{bmatrix} 0.101 & \\ & 0.009 \end{bmatrix}$ | | $\Sigma_t =$ | $\begin{bmatrix} 0.109 & \\ & 0.082 \end{bmatrix}$ | |
| $\Sigma_r =$ | $\begin{bmatrix} 0.000 & 0.097 \\ 0.097 & 0.017 \end{bmatrix}$ | | $\Sigma_r =$ | $\begin{bmatrix} 0.051 & 0.008 \\ 0.008 & 0.000 \end{bmatrix}$ | | $\Sigma_r =$ | $\begin{bmatrix} 0.013 & -0.002 \\ -0.002 & 0.000 \end{bmatrix}$ | |
| $\Sigma_{pt} =$ | $\begin{bmatrix} 0.286 & \\ & 0.543 \end{bmatrix}$ | | $\Sigma_{pt} =$ | $\begin{bmatrix} 0.166 & \\ & 0.608 \end{bmatrix}$ | | $\Sigma_{pt} =$ | $\begin{bmatrix} 0.208 & \\ & 0.552 \end{bmatrix}$ | |
| $\Sigma_{pr} =$ | $\begin{bmatrix} 0.080 & 0.020 \\ 0.020 & 0.066 \end{bmatrix}$ | | $\Sigma_{pr} =$ | $\begin{bmatrix} 0.083 & 0.013 \\ 0.013 & 0.028 \end{bmatrix}$ | | $\Sigma_{pr} =$ | $\begin{bmatrix} 0.070 & 0.023 \\ 0.023 & 0.000 \end{bmatrix}$ | |
| $\Sigma_{tr} =$ | $\begin{bmatrix} 0.019 & \\ & 0.000 \end{bmatrix}$ | | $\Sigma_{tr} =$ | $\begin{bmatrix} 0.002 & \\ & 0.019 \end{bmatrix}$ | | $\Sigma_{tr} =$ | $\begin{bmatrix} 0.000 & \\ & 0.029 \end{bmatrix}$ | |
| $\Sigma_{ptr} =$ | $\begin{bmatrix} 0.106 & \\ & 0.137 \end{bmatrix}$ | | $\Sigma_{ptr} =$ | $\begin{bmatrix} 0.082 & \\ & 0.106 \end{bmatrix}$ | | $\Sigma_{ptr} =$ | $\begin{bmatrix} 0.092 & \\ & 0.129 \end{bmatrix}$ | |

Table 3

*G-Study Variance and Covariance Components: AI Rater*

| Task Completion | | Delivery | | Language Use | |
|---|---|---|---|---|---|
| **SN** | **ER** | **SN** | **ER** | **SN** | **ER** |
| $\Sigma_p = \begin{bmatrix} 0.363 & 0.376 \\ 0.376 & 0.426 \end{bmatrix}$ | | $\Sigma_p = \begin{bmatrix} 0.338 & 0.287 \\ 0.287 & 0.227 \end{bmatrix}$ | | $\Sigma_p = \begin{bmatrix} 0.289 & 0.243 \\ 0.243 & 0.256 \end{bmatrix}$ | |
| $\Sigma_t = \begin{bmatrix} 0.026 & \\ & 0.000 \end{bmatrix}$ | | $\Sigma_t = \begin{bmatrix} 0.004 & \\ & 0.000 \end{bmatrix}$ | | $\Sigma_t = \begin{bmatrix} 0.043 & \\ & 0.000 \end{bmatrix}$ | |
| $\Sigma_r = \begin{bmatrix} 0.169 & 0.005 \\ 0.005 & 0.015 \end{bmatrix}$ | | $\Sigma_r = \begin{bmatrix} 0.044 & 0.022 \\ 0.022 & 0.017 \end{bmatrix}$ | | $\Sigma_r = \begin{bmatrix} 0.061 & 0.056 \\ 0.056 & 0.008 \end{bmatrix}$ | |
| $\Sigma_{pt} = \begin{bmatrix} 0.141 & \\ & 0.462 \end{bmatrix}$ | | $\Sigma_{pt} = \begin{bmatrix} 0.106 & \\ & 0.404 \end{bmatrix}$ | | $\Sigma_{pt} = \begin{bmatrix} 0.109 & \\ & 0.353 \end{bmatrix}$ | |
| $\Sigma_{pr} = \begin{bmatrix} 0.139 & 0.041 \\ 0.041 & 0.005 \end{bmatrix}$ | | $\Sigma_{pr} = \begin{bmatrix} 0.090 & 0.055 \\ 0.055 & 0.030 \end{bmatrix}$ | | $\Sigma_{pr} = \begin{bmatrix} 0.085 & 0.032 \\ 0.032 & 0.000 \end{bmatrix}$ | |
| $\Sigma_{tr} = \begin{bmatrix} 0.050 & \\ & 0.087 \end{bmatrix}$ | | $\Sigma_{tr} = \begin{bmatrix} 0.068 & \\ & 0.167 \end{bmatrix}$ | | $\Sigma_{tr} = \begin{bmatrix} 0.042 & \\ & 0.218 \end{bmatrix}$ | |
| $\Sigma_{ptr} = \begin{bmatrix} 0.314 & \\ & 0.400 \end{bmatrix}$ | | $\Sigma_{ptr} = \begin{bmatrix} 0.315 & \\ & 0.388 \end{bmatrix}$ | | $\Sigma_{ptr} = \begin{bmatrix} 0.376 & \\ & 0.398 \end{bmatrix}$ | |

Similar to the previous research questions, D study results are reported in terms of generalizability coefficients and are presented in Figure 4. Each row corresponds to one of the three domain scores. Overall, the results across the three domain scores were quite similar, with TC yielding slightly higher coefficients than DL and LU, though the differences were minimal. Across all three domain scores, increasing the number of raters had a smaller impact on reliability for human raters compared to AI raters.

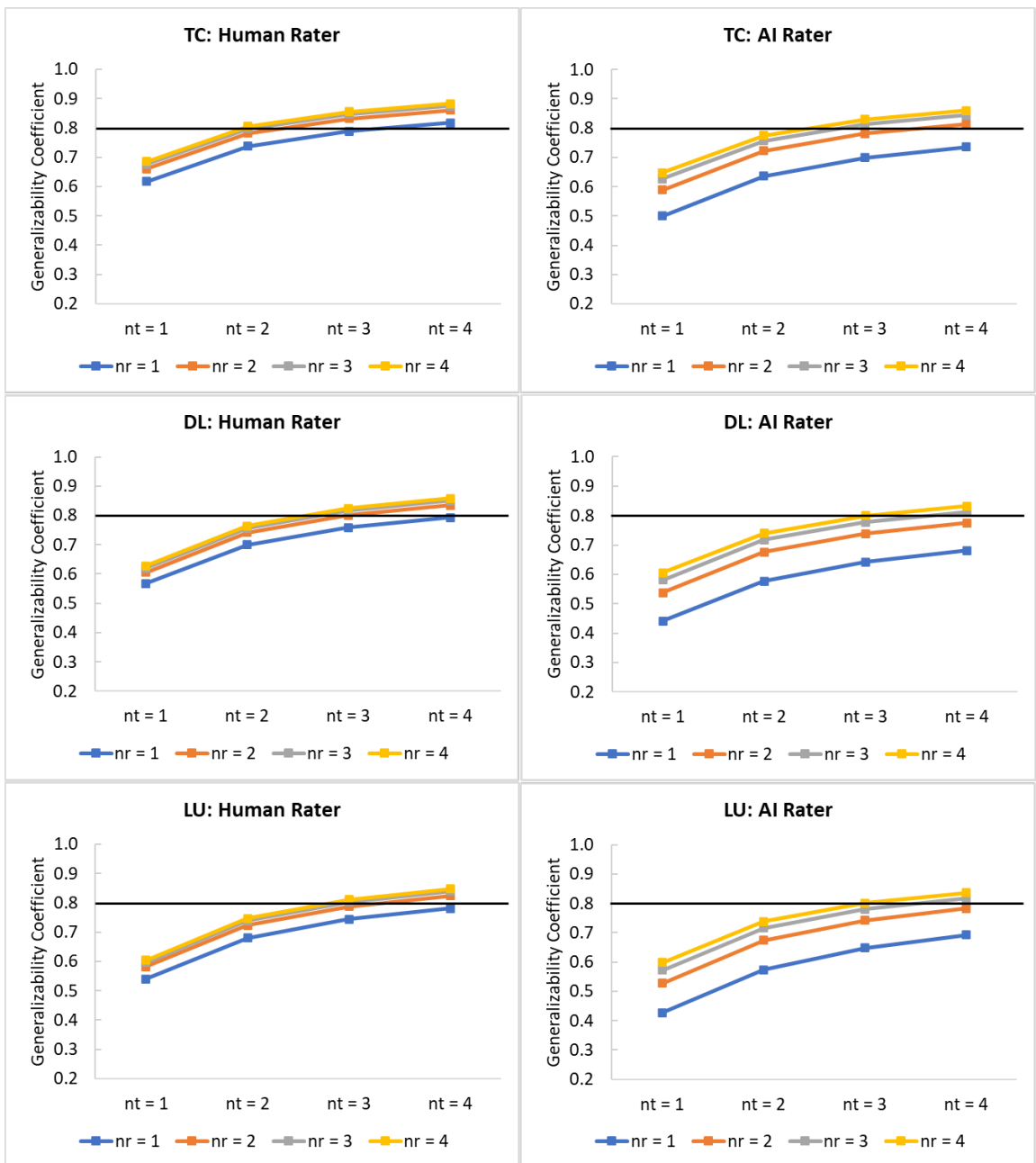

*Figure 4.* Generalizability coefficients: Comparison across domain-specific scores.

## Mixed Use of Human and AI Raters (Research Question 4)

The last research question is concerned about the use of both human and AI raters for scoring. To address this question, a multivariate $p^{\bullet} \times r^{\circ} \times t^{\bullet}$ design was employed, in which the composite mean score over the two rater types was the primary score of interest. Only the overall holistic scores are considered for this research question. A separate analysis was conducted for each of SN and ER task types.

Figure 5 presents the D study results for various sample-size combinations of human and AI raters, with the number of prompts set to either 1 or 2. The left plot corresponds to SN, while the right plot represents ER. Each of the four lines in each plot depicts a specific combination of rater numbers, where the first number indicates the count of human raters and the second number represents the count of AI raters.

An interesting pattern emerged. For both SN and ER, the 1 & 2 combination exhibited higher reliability than the 0 & 3 combination—note that the total number of raters is the same between the two conditions. This interaction was even more pronounced for ER. These findings suggest that incorporating at least one human rater is beneficial.

Another interesting result emerged for ER when comparing the 1 & 1 and 1 & 2 combinations. Surprisingly, the generalizability coefficient for the 1 & 2 condition was lower than that for the 1 & 1 condition, suggesting that adding an AI rater reduced reliability. While this may seem counterintuitive, it occurred due to the weighting used to compute composite score reliability. In the 1 & 1 condition, equal weights of 0.5 were assigned to human and AI raters, whereas in the 1 & 2 condition, the weights were 0.33 and 0.67, respectively, giving twice as much weight to AI raters. Although alternative weighting schemes could be used, proportional weighting based on sample sizes is a common practice.

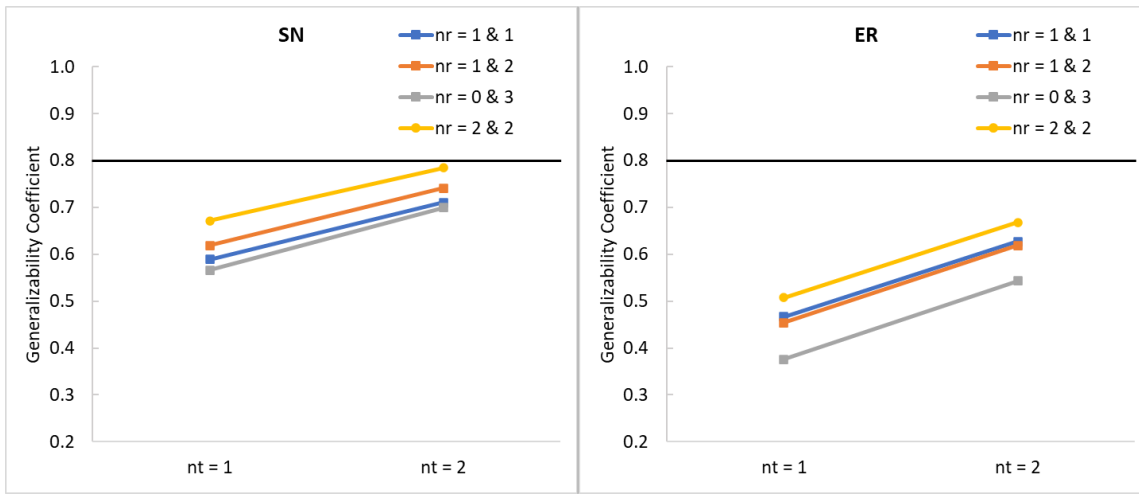

*Figure 5.* Generalizability coefficients: Mixed use of human and AI raters.

### *Inter-rater Reliability*

Often, an inter-rater coefficient is used as a reliability index for ratings. This coefficient is typically computed using the Pearson correlation, based on examinees' scores on a single prompt and ratings from two raters. This type of analysis can be conducted with the current data by using one prompt at a time and the ratings from the two human raters. For AI raters, all possible pairs of raters can be employed, and an average inter-rater coefficient can be calculated.

From the perspective of generalizability theory, the inter-rater coefficient is conceptually associated with the univariate G study $p \times t \times r$ design with two raters evaluating the same prompt. The inter-rater coefficient is equivalent to using a *random $n'_r = 1$* rater in the D study because the inter-rater coefficient, as a correlation, represents reliability for a single rater. Moreover, the D study involves a single prompt, $n'_t = 1$, with the prompt facet being *fixed*—the inter-rater coefficient is computed using a single fixed prompt.

A univariate $p \times t \times r$ design analysis was conducted separately for human and AI raters to obtain G study variance components. Then, the D study sample sizes of $n'_r = 1$ and $n'_t = 1$, with $r$ random and $t$ fixed, were used to compute the generalizability coefficient as an estimate of the inter-rater coefficient. The results are summarized in the top part of Table 4. Each inter-rater coefficient can be interpreted as an average correlation between two raters on each prompt, where the average is taken over all possible pairs of raters and different prompts.

The inter-rater coefficient is usually large, primarily because the prompt facet is fixed. By fixing a facet, which limits generalization, the inter-rater coefficient becomes large because the variance component for the $pt$ interaction (which is typically large) is removed from the error variance and becomes part of the universe score variance. However, as most investigators are interested in generalizing over different prompts in the universe, a more desirable coefficient would be based on the D study treating both raters ($r$) and tasks ($t$) as random. Using both facets as random, the generalizability coefficients using $n'_r = 1$ and $n'_t = 1$ are provided in the bottom part of Table 4. These coefficients are notably lower than the traditional inter-rater coefficients, as they account for variability across both raters and tasks.

Table 4

*Estimated Inter-rater Reliability Coefficients: Overall Scores*

| Human Rater | | AI Rater | |
|:---:|:---:|:---:|:---:|
| $n'_r = 1$ & $n'_t = 1$: $r$ random and $t$ fixed | | | |
| **SN** | **ER** | **SN** | **ER** |
| .750 | .850 | .540 | .651 |
| $n'_r = 1$ and $n'_t = 1$: Both $r$ and $t$ random | | | |
| **SN** | **ER** | **SN** | **ER** |
| .566 | .535 | .392 | .289 |

**Conclusions and Discussion**

This study investigated the reliability of LLMs in scoring AP Chinese writing tasks using generalizability theory. By analyzing scores assigned by both human and AI raters on four tasks (two ER and two SN prompts), this study evaluated the consistency and domain-specific performance of LLMs relative to trained human raters. Findings demonstrated that although LLMs approximated human scoring patterns to some extent, they consistently achieved lower reliability, especially when fewer raters or tasks were involved. Human raters generally yielded higher generalizability coefficients than AI raters across both holistic and analytic scoring dimensions. However, composite scoring involving a combination of human and AI raters improved reliability in specific conditions, which highlights the potential for hybrid scoring approaches in future AES.

This study's findings reveal several critical insights into LLM performance in automated essay scoring for high-stakes language assessments. Firstly, AI raters exhibited considerable variation in ranking examinees relative to task difficulty, which leads to reduced score reliability compared to humans. This finding supports previous concerns that AI raters may struggle with the complexities involved in language assessment tasks. Secondly, reliability differed notably by task type: SN tasks demonstrated higher score consistency compared to ER tasks. The structured nature of SN, aided by visual prompts, appeared more manageable for AI interpretation. In contrast, the ER tasks might be more challenging to score, as they required understanding subtle, real-world context and pragmatic language use. Thirdly, across the scoring domains of task completion, delivery, and language use, the task completion domain exhibited slightly higher reliability. This may be due to its more objective evaluation criteria, in contrast to the more subjective judgments required for delivery and language use. Notably, AI raters showed greater variability when scoring these subtle aspects of language, which suggests they might struggle with making complex linguistic judgments. Lastly, the composite scoring analysis provided valuable insights. Employing at least one human rater alongside AI improved scoring reliability. However, increasing the number of AI raters did not consistently enhance reliability and, in some cases, it lowered it. This underscores the necessity for balanced integration of AI and human evaluations. Overall, these results suggest that carefully structured hybrid models, rather than full reliance on LLMs alone, offer a practical and effective approach for improving reliability in large-scale writing assessments.

There are a few limitations in this research. First, the study included a relatively small sample size (30 students producing 120 essays). While generalizability theory is robust for estimating variance components even with smaller samples, applying these findings to broader student populations remains constrained. Future research can replicate the current approach using larger and more diverse participant groups to ensure broader generalizability.

Second, the LLMs employed were not specifically fine-tuned for the AP Chinese scoring tasks but rather guided through structured prompts based on official rubrics. Although this reflects real-world educational settings, it may have limited the highest level of reliability that could be achieved. Future research could explore whether fine-tuning or reinforcement learning might help reduce the reliability gap between AI and human raters.

Third, although the research controlled for rater training and task type, factors such as students' linguistic backgrounds, familiarity with task types, and individual writing strategies were not explicitly addressed. These elements potentially introduced additional variability that could influence scoring outcomes. Future investigations could also consider how LLMs perform across varying learner profiles and contexts.

Lastly, the study focused solely on AP Chinese writing assessments, characterized by unique cultural and linguistic expectations. In this case, the findings may not be directly generalizable to other language assessments involving different writing genres, such as argumentative or analytical essays. Broader comparative research across diverse language assessments and content areas is essential to better understand how well LLM-based scoring models work and where they can be effectively applied.

**Appendix A.**

**AI Training Protocol for Grading SN2 Student Samples**

Based on this prompt, grading rubric, and student writing samples, grade student's writing sample.

Prompt: In this task, you will be asked to write in Chinese for a specific purpose and to a specific person. You should write in as complete and culturally appropriate a manner as possible, taking into account the purpose and the person described. The four pictures present a story. Imagine you are writing the story to a friend. Narrate a complete story as suggested by the pictures. Give your story a beginning, a middle, and an end.

**The first picture:** A woman is at the airport, standing in front of the flight information board. She is also looking at her phone to check the details.

**The second picture:** The woman stops by a restaurant or shop to get some fast food. She leaves her phone on the table.

**The third picture:** The woman goes to check in for her flight. While standing in line, she realizes that her phone is missing.

**The fourth picture:** The woman runs back to the restaurant or shop, and an employee returns her phone to her.

The task for you is to rate the student responses. Assess each response holistically based on three key criteria: task completion, delivery, and language use, with task completion being the most critical factor. Start by determining whether the student fully addresses all aspects of the four pictures. Then, evaluate delivery and language use before assigning an overall score. Additionally, please provide three analytic scores—one for each criterion—following the rubric's guidelines. When rating, refer to the sample responses for each score level as benchmarks. In total, provide four scores: one overall score and three analytic scores. Tips: Student responses will be graded based on task completion, delivery, and language use. However, it is important to apply a holistic grading approach. Among these criteria, task completion is the most important. If a student addresses only three pictures, they will receive 3 points, regardless of the quality of their language use and delivery. A student will receive 4 points if they discuss all four pictures. In this context, use grading rubrics 5 and 6 as references for evaluating language use and delivery. If a student addresses three pictures but demonstrates weak language use and delivery, they may receive 2 points. A student who addresses only one or two pictures but includes one or two sentences may also receive 2 points. If a student writes only isolated words related to the prompts (pictures), they may receive 1 point.

**Grading Rubric:**

Score of 0: UNACCEPTABLE: 1) Contains nothing that earns credit 2) Completely irrelevant to the stimulus 3) Not in Chinese characters, NR (No Response), BLANK (no response)

Samples:

Score 0:
再见

Score 0:
I don't know.

Score of 1: Very Weak: Demonstrates lack of competence in interpersonal writing. - TASK COMPLETION: 1) E-mail addresses stimulus only minimally 2) Lacks organization and coherence; very disjointed sentences or isolated words -DELIVERY: 1) Constant use of register inappropriate to situation -LANGUAGE USE: 1) Insufficient, inappropriate vocabulary, with frequent errors that significantly obscure meaning; constant interference from another language 2) Little or no control of grammatical structures, with frequent errors that significantly obscure meaning

Score 1:
小红在机场跑步。

Score 1:
丽丽觉得饿，要有吃。

Score of 2: Weak: Suggests lack of competence in interpersonal writing - TASK COMPLETION: 1) E-mail addresses topic only marginally or addresses only some aspects of stimulus 2) Scattered information generally lacks organization and coherence; minimal or no use of transitional elements and cohesive devices; fragmented sentences -DELIVERY: 1) Frequent use of register inappropriate to situation -LANGUAGE USE: 1) Minimal appropriate vocabulary, with frequent errors that obscure meaning; repeated interference from another language 2) Limited grammatical structures, with frequent errors that obscure meaning

Score 2:
有一天早上七点半卧在床和我去了飞机窗。在飞机窗我用我的手机，因为我可以看一个餐厅。我社非常饿。我去了一个很好的西班牙的饭店。我说："我可以买了一个科拉和一个 taco。"如果我忘记了我的手机。

Score 2:
我的朋友蒂芙尼乘飞机飞往墨西哥. 但她饿了，她想吃点饭。

Score of 3: Adequate Suggests competence in interpersonal writing -TASK COMPLETION: 1) E-mail addresses topic directly but may not address all aspects of stimulus 2) Portions may lack organization or coherence; infrequent use of transitional elements and cohesive devices; disconnected sentences -DELIVERY: 1) Use of register appropriate to situation is inconsistent or includes many errors -LANGUAGE USE: 1) Limited appropriate vocabulary and idioms, with frequent errors that sometimes obscure meaning; intermittent interference from another language 2) Mostly simple grammatical structures, with frequent errors that sometimes obscure meaning

Score 3:
丽丽在飞机站的时候看一看她的手机有没有时间可以那一些零食。他看到有四十分钟，决定会拿一个汉堡包所以在飞机感觉饿的时候能吃食物。他在去飞机排的时候意识到他在快餐店忘了她的手机。她很快跑步去哪里拿她的手机，好性服务员给了丽丽的手机。他感觉很意义因为他的压力很高而且没有那么多钱能买个手机。

Score 3:
今天李天说去机场。她去加拿大旅行。他登上飞机的时候想买一点水和小吃。上飞机的时候，他喜欢看电影和吃很多小吃。买玩的时候，她回去上飞机。但是她在那个小饭店忘了手机。所以她再去找她的手机。然后上飞机了。

Score of 4: Good Demonstrates competence in interpersonal writing -TASK COMPLETION: 1) E-mail addresses all aspects of stimulus but may lack detail or elaboration 2) Generally organized and coherent; use of transitional elements and cohesive devices may be inconsistent; discourse of paragraph length, although sentences may be loosely connected -DELIVERY: 1) May include several lapses in otherwise consistent use of register appropriate to situation -LANGUAGE USE: 1) Mostly appropriate vocabulary and idioms, with errors that do not generally obscure meaning 2) Mostly appropriate grammatical structures, with errors that do not generally obscure meaning

Score 4:
今天是小美第一次坐飞机去中国北京。小美到飞机场检查她的手机先看她的飞机票。她找到她要去的门儿以后感觉很饿，所以小美去找一家饼店。她点饭的时候留他的手机在桌子上，又饿又着急小美忘了拿她的手机。小美赶快回去找她的门儿之后，突然她发现她忘了她的手机在饼店。她速度快跑去接她的手机。

Score 4:
有一天有一位小姐去機場。她到機場的時候走到螢幕去看一次找她的航空。忽然，她感覺有點餓，所以她決定去墨西哥餐館點菜。他考慮買什麼菜吃的時候把手機拉在餐館櫃檯上。她買了一些飲食和一杯可樂。過一陣子她就去她的航班的登機門。在登機門馬上要上飛機的時候，她發現她的手機不在身邊。想來想去決定她把她的手機拉在墨西哥餐館裏。她突然跑到墨西哥餐館去。服務員認識了她所以把手機還給她。

Score of 5: Very good Suggests excellence in interpersonal writing -TASK COMPLETION: 1) E-mail addresses all aspects of stimulus 2) Well organized and coherent, with a progression of ideas that is generally clear; some use of transitional elements and cohesive devices; connected discourse of paragraph length -DELIVERY: 1) Consistent use of register appropriate to situation except for occasional lapses -LANGUAGE USE: 1) Appropriate vocabulary and idioms, with sporadic errors 2) Variety of grammatical structures, with sporadic errors

Score 5:
今天安娜打算坐飞机回中国。在机场，当她用她的手机准备登机的时候，她有一点饿。于是，她在机场她找一家餐厅，想买一些辣的食物。她找到了一家餐厅，但是她还没想好到底要吃什么。当她思考的时候，她把手机放了桌子上，点完饭以后她就很快地跑去登机口，生怕自己误了飞机。突然，她意识到自己把手机忘在了餐厅，于是回去取，热心的服务员把手机给她。

Score 5:
小小在机场等飞机。她看到自己的航班还没有到达，觉得又饿又无聊，于是决定去买点吃的。她走到一家墨西哥餐厅，把手机放在柜台上，然后点了几样她喜欢的菜。买完食物后，小小开心地继续等她的航班。后来，当开始登机时，小小突然发现她的手机不见了！她立刻想起来手机还放在餐厅的柜台上。于是，她赶紧跑回那家墨西哥餐厅，工作人员正好把手机保管好，还给了她。小小松了一口气，拿着手机顺利登上了飞机。

Score of 6: Excellent Demonstrates excellence in interpersonal writing -TASK COMPLETION: 1) E-mail addresses all aspects of stimulus with thoroughness and detail 2) Well organized and coherent, with a clear progression of ideas; use of appropriate transitional elements and cohesive devices; well-connected discourse of paragraph length -DELIVERY: 1) Consistent use of register appropriate to situation -LANGUAGE USE: 1) Rich and appropriate vocabulary and idioms, with minimal errors 2) Wide range of grammatical structures, with minimal errors

Score 6:
有一天，李美决定了他想坐飞机去日本。她觉得日本很漂亮，她也有很多朋友住在日本。所以他买了飞机票，去飞机场。李美到飞机场的时候，她用她的手机看几点她会坐飞机。她看到了她还有一个小时，所以去一个小餐馆叫 Cafe Mexico 买午饭。她饿死了，告诉餐馆的男人，『请你等一下，我会想想什麽买。』但是李美想的时候，她放手机在前面的桌子。李美点她的菜。她吃完以后，去 Gate A8，因为她会做的飞机在这裡。可要李美会坐飞机的时候，突然想想，『我的手机在哪裡 ？』两分以后，她觉得她知道手机在哪裡。李美快地跑步回去 Cafe Mexico。她到了 Cafe Mexico 的时候，餐馆的男人给她她的手机。李美说『谢谢你，谢谢你，我真的丢三落四的，对不起。』男人说『哈哈，没问题。』李美拿著她的手机，很高兴，因为这样她可以坐飞机去日本。

Score 6:
有一位女孩前往机场，准备搭乘航班出发。当她在候机时，感到十分饥饿，便决定前往附近的一家餐馆买些食物充饥。点餐时，她一时疏忽，把手机放在了柜台上，随后便专心挑选餐点。买完餐以后，她匆匆前往登机口准备登机。就在即将排队登机的那一刻，她突然意识到手机不见了！她感到十分慌张，经过短暂思索，她迅速意识到可能把手机遗忘在餐馆里了。她立刻飞奔回去，希望还能找回她的手机。幸运的是，手机依然安然无恙地放在柜台上，餐馆的服务员早已将它妥善保管。看到女孩回来，服务员微笑着把手机递给她。她连声道谢："真的非常感谢你！幸好有你，不然我可能就赶不上飞机了。"服务员笑着回答："别客气，祝你旅途顺利！"女孩拿回手机，长舒了一口气，带着感激和喜悦的心情，顺利前往登机口，踏上了她的旅程。

**Appendix B.**

Table 1

*Mean and Standard Deviation of Human and AI Rater Scores Across Tasks and Scoring Domains*

| Task | Rater | Overall | | Task Completion | | Delivery | | Language Use | |
|------|-------|------|------|------|------|------|------|------|------|
| | | Mean | SD | Mean | SD | Mean | SD | Mean | SD |
| | H1 | 3.69 | 0.76 | 4.17 | 1.04 | 3.76 | 0.87 | 3.83 | 0.89 |
| | H2 | 4.07 | 0.75 | 4 | 0.85 | 4 | 0.85 | 3.93 | 0.92 |
| | A1 | 3.48 | 0.99 | 3.66 | 0.77 | 3.38 | 1.05 | 3.41 | 1.02 |
| | A2 | 2.97 | 0.82 | 3.24 | 0.74 | 2.97 | 0.82 | 2.97 | 0.82 |
| SN1 | A3 | 3.69 | 1.17 | 4.17 | 1.23 | 3.45 | 1.12 | 2.93 | 1.03 |
| | A4 | 2.79 | 0.77 | 2.79 | 0.77 | 2.62 | 0.86 | 2.72 | 0.96 |
| | A5 | 3.76 | 1.09 | 3.76 | 1.09 | 3.76 | 1.09 | 3.76 | 1.09 |
| | A6 | 3.1 | 0.77 | 3.17 | 0.85 | 2.86 | 0.79 | 3.07 | 0.75 |
| | A7 | 2.45 | 0.83 | 2.83 | 0.71 | 2.24 | 0.79 | 2.24 | 0.79 |
| | | Overall | | Task Completion | | Delivery | | Language Use | |
| | | Mean | SD | Mean | SD | Mean | SD | Mean | SD |

| | | Overall | | Task Completion | | Delivery | | Language Use | |
|---|---|---|---|---|---|---|---|---|---|
| | | Mean | SD | Mean | SD | Mean | SD | Mean | SD |
| ER1 | H1 | 3.28 | 0.88 | 3.9 | 1.21 | 3.52 | 0.87 | 3.41 | 0.78 |
| | H2 | 3.83 | 0.97 | 3.69 | 1.07 | 3.72 | 1 | 3.59 | 0.91 |
| | A1 | 4.24 | 1.46 | 4.1 | 1.52 | 4.24 | 1.38 | 4.28 | 1.39 |
| | A2 | 3.41 | 0.68 | 3.86 | 0.83 | 3.86 | 0.92 | 3.34 | 0.72 |
| | A3 | 2.9 | 0.9 | 3.28 | 1.19 | 2.9 | 0.82 | 2.59 | 0.95 |
| | A4 | 3.45 | 0.91 | 3.34 | 1.11 | 3.55 | 0.78 | 3.34 | 0.9 |
| | A5 | 3.48 | 1.24 | 3.55 | 1.33 | 3.93 | 1.16 | 3.21 | 1.01 |
| | A6 | 2.9 | 0.77 | 3.17 | 0.85 | 2.97 | 0.78 | 2.62 | 0.56 |
| | A7 | 2.83 | 0.76 | 3.24 | 0.91 | 2.59 | 0.91 | 2.38 | 0.73 |

| | | Overall | | Task Completion | | Delivery | | Language Use | |
|---|---|---|---|---|---|---|---|---|---|
| | | Mean | SD | Mean | SD | Mean | SD | Mean | SD |
| SN2 | H1 | 3.24 | 0.69 | 3.62 | 0.9 | 3.24 | 0.64 | 3.31 | 0.6 |
| | H2 | 3.69 | 0.66 | 3.72 | 0.8 | 3.62 | 0.68 | 3.52 | 0.69 |
| | A1 | 3.48 | 0.95 | 3.52 | 0.83 | 3.41 | 0.87 | 3.41 | 0.87 |
| | A2 | 2.93 | 1.03 | 3.31 | 0.93 | 3.07 | 0.92 | 2.86 | 0.99 |
| | A3 | 2.69 | 0.85 | 3.45 | 1.09 | 2.69 | 0.85 | 2.38 | 0.82 |
| | A4 | 2.9 | 0.67 | 2.9 | 0.67 | 2.9 | 0.67 | 2.9 | 0.67 |
| | A5 | 3.07 | 1.07 | 3.03 | 1.05 | 3.07 | 1.07 | 3.03 | 1.05 |
| | A6 | 2.86 | 0.95 | 3.03 | 1.18 | 2.83 | 0.66 | 2.79 | 0.77 |
| | A7 | 2.72 | 0.75 | 3.41 | 0.82 | 2.9 | 0.67 | 2.41 | 0.68 |

| | | Overall | | Task Completion | | Delivery | | Language Use | |
|---|---|---|---|---|---|---|---|---|---|
| | | Mean | SD | Mean | SD | Mean | SD | Mean | SD |
| ER2 | H1 | 3.72 | 1.13 | 4 | 1.41 | 3.93 | 1.36 | 4.07 | 1.31 |
| | H2 | 4.03 | 1.18 | 3.83 | 1.28 | 3.83 | 1.28 | 3.86 | 1.3 |
| | A1 | 3.07 | 0.96 | 3.21 | 1.15 | 3.14 | 0.99 | 3.14 | 0.88 |

| A2 | 3.9 | 1.23 | 3.9 | 1.23 | 4 | 1.07 | 3.86 | 1.25 |
|----|------|------|------|------|------|------|------|------|
| A3 | 3.72 | 1.25 | 3.83 | 1.26 | 3.83 | 1.23 | 3.72 | 1 |
| A4 | 3.59 | 0.68 | 3.52 | 0.69 | 3.59 | 0.78 | 3.48 | 0.74 |
| A5 | 3.83 | 1.39 | 3.55 | 1.27 | 3.66 | 1.23 | 3.62 | 1.24 |
| A6 | 3.07 | 1.03 | 3.03 | 1.02 | 3.03 | 1.02 | 3 | 1 |
| A7 | 3.34 | 0.9 | 3.34 | 1.01 | 3.38 | 0.82 | 3.17 | 0.93 |